\documentclass[conference]{IEEEtran}
\IEEEoverridecommandlockouts
\usepackage{cite}
\usepackage{amsmath,amssymb,amsfonts}
\usepackage{algorithmic}
\usepackage{graphicx}
\usepackage{textcomp}
\usepackage{xcolor}
\usepackage{hyperref}
\usepackage{float}
\usepackage{cite}
\def\BibTeX{{\rm B\kern-.05em{\sc i\kern-.025em b}\kern-.08em
    T\kern-.1667em\lower.7ex\hbox{E}\kern-.125emX}}
    
\begin{document}

\title{Evaluating Low-Resource Lane Following Algorithms for Compute-Constrained Automated Vehicles\\
%{\footnotesize \textsuperscript{*}Note: Sub-%titles are not captured in Xplore and
%should not be used}
\thanks{This project is supported by the National Science Foundation REU Site Awards \#2150096 and \#2150292. The electric vehicles used for this research are sponsored by Hyundai MOBIS, US Army GVSC, NDIA, DENSO, SoarTech, Realtime Technologies, Veoneer, Dataspeed, GLS\&T, and LTU.}
}

\author{
\IEEEauthorblockN{Beñat Froemming-Aldanondo*}
\IEEEauthorblockA{\textit{Dept. of Computer Science \& Engineering} \\
\textit{University of Minnesota}\\
Minneapolis, United States \\
froem076@umn.edu}
*Corresponding author
~\\
\and
\IEEEauthorblockN{Tatiana Rastoskueva}
\IEEEauthorblockA{\textit{Dept. of Computer Science} \\
\textit{The University of Arizona}\\
Tucson, United States \\
trastoskueva@arizona.edu}
~\\
\and
\IEEEauthorblockN{Michael Evans}
\IEEEauthorblockA{\textit{Dept. of Computer Science} \\
\textit{Old Dominion University}\\
Norfolk, United States \\
mevan028@odu.edu}
~\\
\and
\IEEEauthorblockN{Marcial Machado}
\IEEEauthorblockA{\textit{Dept. of Computer Science \& Engineering} \\
\textit{The Ohio State University}\\
Columbus, United States \\
marcialmachado0522@gmail.com}
~\\
\and
\IEEEauthorblockN{Anna Vadella}
\IEEEauthorblockA{\textit{Dept. of Computer Science \& SWE} \\
\textit{Butler University}\\
Indianapolis, United States \\
avadella@butler.edu}
~\\
\and
\IEEEauthorblockN{Rickey Johnson}
\IEEEauthorblockA{\textit{Dept. of Computer Science} \\
\textit{North Carolina A\&T State University}\\
Greensboro, United States \\
rdjohnson2027@gmail.com}
~\\
\and
\IEEEauthorblockN{Luis Escamilla}
\IEEEauthorblockA{\textit{Dept. of Computer Science} \\
\textit{New Mexico State University}\\
Las Cruces, United States \\
loktavio101@gmail.com}
~\\
\and
\IEEEauthorblockN{Milan Jostes, Devson Butani, Ryan Kaddis, Chan-Jin Chung}
\IEEEauthorblockA{\textit{Dept. of Math \& Computer Science} \\
\textit{Lawrence Technological University}\\
Southfield, United States \\
\{mjostes, dbutani, rkaddis, cchung\}@ltu.edu}
~\\
\and
\IEEEauthorblockN{Josh Siegel}
\IEEEauthorblockA{\textit{DeepTech Lab} \\
\textit{Michigan State University}\\
East Lansing, United States \\
jsiegel@msu.edu}
}
\maketitle

\begin{abstract}
Reliable lane-following is essential for automated and assisted driving, yet existing solutions often rely on models that require extensive computational resources, limiting their deployment in compute-constrained vehicles. We evaluate five low-resource lane-following algorithms designed for real-time operation on vehicles with limited computing resources. Performance was assessed through simulation and deployment on real drive-by-wire electric vehicles, with evaluation metrics including reliability, comfort, speed, and adaptability.  The top-performing methods used unsupervised learning to detect and separate lane lines with processing time under 10 ms per frame, outperforming compute-intensive and poor generalizing deep learning approaches. These approaches demonstrated robustness across lighting conditions, road textures, and lane geometries. The findings highlight the potential for efficient lane detection approaches to enhance the accessibility and reliability of autonomous vehicle technologies. Reducing computing requirements enables lane keeping to be widely deployed in vehicles as part of lower-level automation, including active safety systems. 
\end{abstract}


\section{Introduction}
Lane following is a fundamental capability for autonomous vehicles, requiring accurate detection and tracking of lane markings to ensure proper road positioning and collision avoidance. This paper presents the implementation, testing, and validation of five low-resource lane-following algorithms on real drive-by-wire vehicles, with the goal of identifying approaches that balance efficiency and performance under constrained computing resources.

Data collection and model training in autonomous driving often demand vast resources and high-performance hardware. According to estimates released by Nvidia in 2017 \cite{grzywaczewski2017training}, a fleet of 100–125 vehicles can generate 203–595 PB of raw data per year, resulting in 104–487 TB after preprocessing. Training deep models such as ResNet-50 \cite{7780459} on these data volumes can take upwards of 113–528 days on a single NVIDIA DGX-1, necessitating 97–1{,}056 machines to achieve a 7-day target. This burden is further compounded by the need for real-world and simulated data to improve generalization.

However, many smaller autonomous platforms (e.g., mobility aids, golf carts, and delivery robots) cannot afford the computing overhead associated with large-scale models such as LaneSegNet \cite{li2024lanesegnetmaplearninglane} or MapTR \cite{liao2023maptrstructuredmodelinglearning}, particularly if they must operate at or above human reaction times. Lightweight algorithms, including traditional computer vision and compact learning-based methods, are more viable in such scenarios. They also serve as valuable backups during extreme conditions (e.g., high temperatures, reduced power) and can complement larger models by detecting anomalies in safety-critical environments.

During previous editions of the Research Experience for Undergraduates (REU) at Lawrence Technological University (LTU), supported by the National Science Foundation (NSF), and run jointly with Michigan State University (MSU), lane detection algorithms were developed and validated on full-scale electric vehicles \cite{9973493}. Those efforts focused on traditional computer vision with OpenCV-based pipelines. Building on these foundations, this paper introduces more efficient machine learning techniques aimed at low-resource environments.

We evaluate five algorithms: (1) Largest White Contour, (2) Lane Line Approximation using Least Squares Regression, (3) Linear Lane Search with K-Means, (4) Lane Line Discrimination using DBSCAN, and (5) DeepLSD Lane Detection. Their performance is quantified using four key metrics: reliability (success rate of completing laps), comfort (smoothness of steering), speed (lap time), and adaptability (robustness to varied driving conditions). Experiments were conducted on a standard laptop to simulate a compute-constrained environment similar to many smaller autonomous vehicles. This paper also covers lane-centering strategies, system architecture, and real-world implementation challenges, concluding with an experimental analysis that identifies the most effective algorithm.

The remainder of this paper is organized as follows: Section II reviews related work; Section III outlines the methodology and resources; Section IV describes the software architecture with four nodes; Section V covers preprocessing;  Section VI details five lane detection algorithms; Section VII discusses vehicle control algorithms; Section VIII presents experimental results; and Section IX concludes with future research directions. The project code is available at 
\url{https://github.com/benatfroemming/REU-2024-Lane-Following}.

\section{Related Work}

Accurate lane detection enables autonomous vehicles to continuously monitor their position and state, making informed decisions for safe driving. Developing robust lane detection algorithms is challenging due to diverse road conditions, varied lane markers, and changing geometry. As a result, extensive research has focused on creating reliable lane detection methods.

\subsection{Traditional Methods}
Traditional approaches typically adopt a vision-based pipeline consisting of image preprocessing, feature extraction, and marker detection. They are computationally light, making them suitable for real-time inference on constrained computing platforms. Many pipelines start by extracting a Region of Interest (ROI) to focus on the road at the bottom of the image \cite{WU20142756}, with the ROI being adaptable to changing environments such as curves \cite{8230303}. Common techniques include Canny edge detection \cite{4767851} and the Hough Transform for line detection \cite{MATAS2000119}, with lines then filtered to match lane line characteristics. Some studies employ the random sample consensus (RANSAC) algorithm in conjunction with the least squares method for estimating lane model parameters based on feature extraction \cite{7098273}. Additionally, clustering techniques like k-means \cite{8515938} and DBSCAN \cite{8407261} are used to separate lane markings. To improve line fitting, a bird's-eye view transformation is often applied, making the image appear as though it was taken from above, which causes the lines to appear parallel rather than converging due to depth \cite{author2024birds}. Some methods also incorporate color-based segmentation \cite{1364007} and morphological operations to enhance lane marker visibility under varying lighting conditions.

Due to the complexity and cost of developing using full-scale autonomous vehicles, most research on lane-following using traditional vision approaches has not been tested outside of a virtual environment. The performance of an algorithm can differ dramatically in a real-world environment, which may involve a dynamic context and confounding factors such as imperfect kinematic modeling. 

\subsection{AI Methods}
AI-driven lane detection typically employs deep learning, exploiting large datasets that sometimes integrate radar, GPS, or LiDAR in addition to camera inputs. In most studies, models are either trained and tested on both public benchmarks and proprietary datasets or solely on custom data.

Three prominent AI strategies include segmentation, anchor-based, and parameter-based methods.

In segmentation approaches, the most common architecture is the Encoder-Decoder architecture. The models take an image as an input, and output another image, where each pixel is classified as part of a lane line or not. The most popular model include LaneNet \cite{wang2018lanenetrealtimelanedetection}, LaneSegNet, and MapTR. However other algorithms have shown good results like, SCNN (Spatial Convolutional Neural Network) \cite{pan2017spatialdeepspatialcnn} which employing spatial CNN layers, RESA (REcurrent Spatial Attention) \cite{zheng2021resarecurrentfeatureshiftaggregator} which leverages spatial attention, and CurveLane-NAS which uses Neural Architecture Search (NAS) \cite{xu2020curvelanenasunifyinglanesensitivearchitecture}. Anchor-based approaches are inspired by object detection, utilizing anchors specifically for lines. The goal is to define these anchors and then compute the deviation of the detected lines from them. LaneATT \cite{tabelini2020eyeslanerealtimeattentionguided}, a state-of-the-art model, employs attention-based anchor generation for this purpose. In contrast, parameter-based approaches directly regress the polynomial equations that describe the lines. Typically, a third-order polynomial (for complex curves) is used, with a fixed number of lines. Models like PolyLaneNet \cite{tabelini2020polylanenetlaneestimationdeep} are leaders in this category.

These models have been successfully deployed in real autonomous vehicles. However, they demand substantial computing power for both training and inference. For example, LaneSegNet was trained on 8 NVIDIA Tesla V100 GPUs and operates at 14.7 FPS on an NVIDIA A100 GPU. LaneNet runs at 330 FPS on a NVIDIA Titan Xp GPU, but only 26 FPS on an NVIDIA Jetson TX1. Similarly, MapTR was trained using 8 NVIDIA GeForce RTX 3090 GPUs, with the nano version running at 25.1 FPS on an RTX 3090. 
Such complex models remain inaccessible for real-time deployment under stringent hardware constraints, especially where continuous operation at or above human reaction time is required.

\section{Material and Methods}
\subsection{Simulation and Real Environment}
Two simulation tools were used to validate the lane-following algorithms before real-world testing: Simple-Sim \cite{simplesim} and Gazelle-Sim \cite{gazellesim}. Both simulators run in ROS, enabling an environment where either a single (Simple-Sim) or multiple (Gazelle-Sim) virtual robots can be controlled. Although these platforms were helpful for initial validation, performance data presented in this paper was collected exclusively from tests conducted on full-scale drive-by-wire vehicles.

\begin{figure}[h]
\centerline{\includegraphics[width=0.85\linewidth]{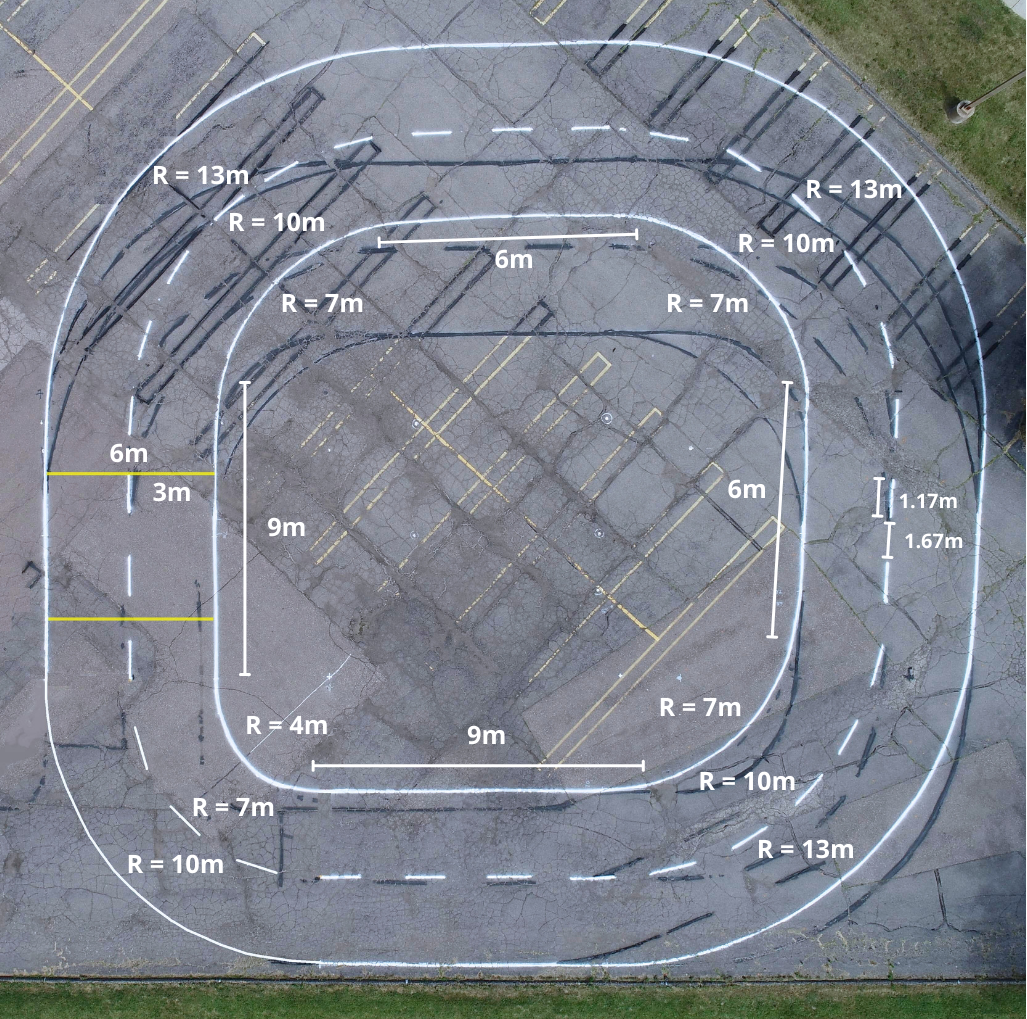}}
\caption{Aerial view of the Lot H course in LTU.}
\label{fig1}
\end{figure}

Following simulation trials, each algorithm was adapted for real-world testing on a dedicated course in Parking Lot~H at Lawrence Technological University, Southfield, Michigan, USA. This circular test course simulates real-world adverse conditions such as potholes, sharp curves, faded and narrow lane markings, cracks, and extraneous lines. It also introduces challenges like tree shadows, sun glare, and puddles. An aerial view of the course is shown in Fig.~\ref{fig1}. The onboard computing platform for each vehicle was an MSI Gaming Laptop featuring an Intel 8-Core i7-11800H processor, 16GB of RAM, a 512GB SSD, and a GeForce RTX 3050 Ti 4GB graphics card.

\subsection{Vehicle Specifications}
\begin{figure}[h]
\centerline{\includegraphics[width=0.95\linewidth]{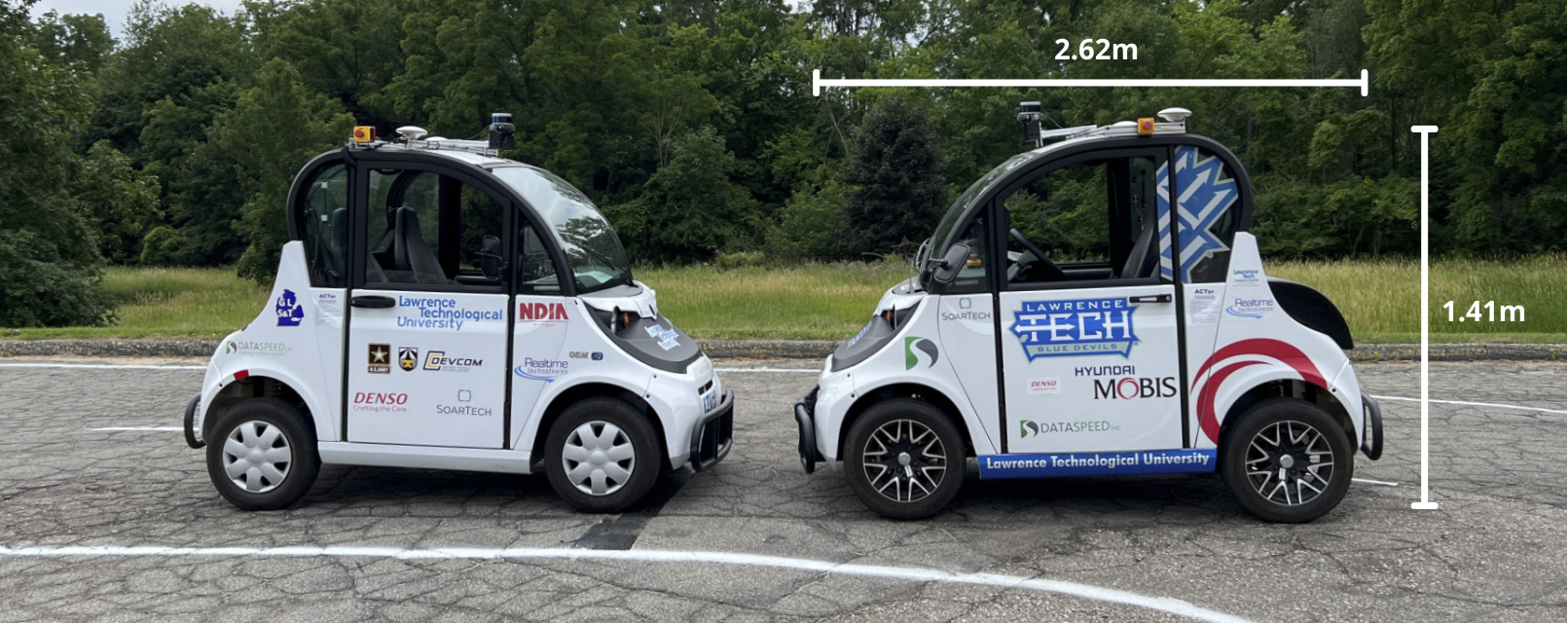}}
\caption{ACTors 1 and 2.}
\label{fig2}
\end{figure}

The two vehicles, referred to as ACTors (\textbf{A}utonomous \textbf{C}ampus \textbf{T}ransp\textbf{or}t) 1 and 2, are modified Polaris Gem e2 models equipped with a Dataspeed Drive-b=-Wire (DBW) kit, high-dynamic-range (HDR) cameras for lane following, 2D and 3D LiDAR sensors, and two Swift Piksi GPS units. Each vehicle is also fitted with a Netgear router, power inverter, and a removable computer system for ROS-based control and networking. The Polaris Gem e2 platform achieves a top speed of 25 miles per hour and a range of approximately 30 miles (Fig~\ref{fig2}).

\section{Architecture and Dynamic Reconfigure}
All lane-following algorithms employ the same ROS-based architecture, designed to be modular and facilitate quick switching between methods and environments. The system also supports parameter tuning via a dynamic reconfigure interface, which allows users to graphically activate the vehicle, adjust the vehicle's speed, tune the algorithms' parameters, and modify the steering sensitivity. The ROS-based architecture consists of 4 nodes: Preprocessor, Lane Detector, Vehicle Controller, and Vehicle Node.

\section{Preprocessing}
The Preprocessor node refines raw camera frames to highlight road markers. First, a median blur is applied to reduce noise while preserving edges. Next, the image is converted to a single-channel grayscale format, facilitating white-pixel thresholding. After thresholding, only the relevant white regions are kept, and the top portion of the frame is cropped to isolate the road region of interest (ROI). The ROI only looks at the road and avoids other extraneous noises like buildings, trees, and the sky. These steps, implemented using OpenCV, are common to all five algorithms. Each can be disabled or tuned using Dynamic Reconfigure, e.g., the upper and lower white threshold values when creating the mask. The processed image is then passed to the lane detection algorithm. 

\section{Lane Detection Algorithms}
This section outlines five lightweight algorithms adapted from the literature for real-time lane detection on resource-constrained vehicles. Most rely on traditional computer vision and unsupervised machine learning to accommodate limited on-board computing power common in automated vehicles.

\subsubsection{Largest White Contour}

\begin{figure}[h]
\centerline{\includegraphics[width=0.55\linewidth]{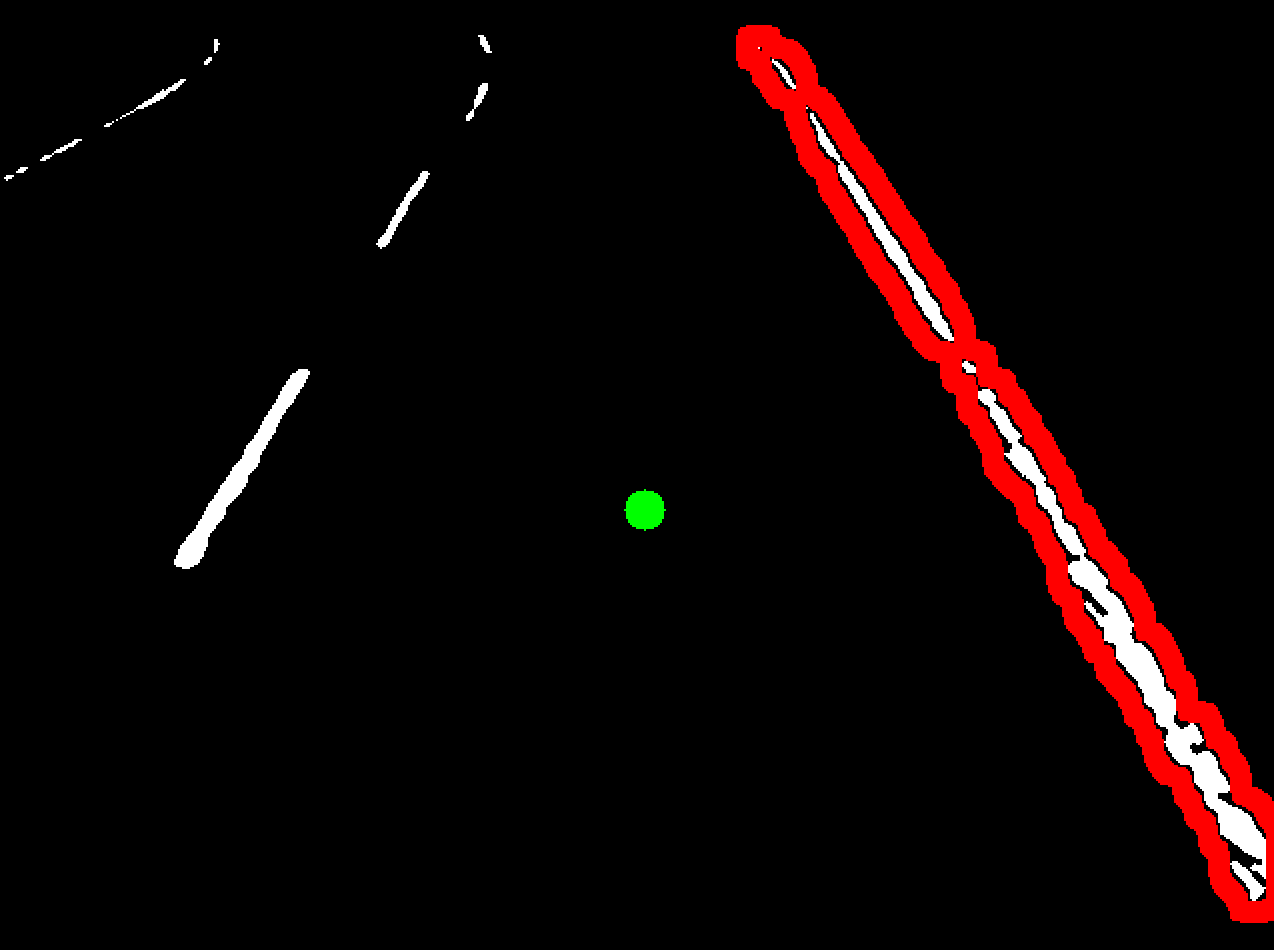}}
\caption{Largest Contour and Offset Point.}
\label{fig3}
\end{figure}

This baseline method identifies and tracks the largest contiguous set of white pixels in the preprocessed image, known as a contour~\cite{chung2022simple}. OpenCV is used to obtain a list of all white contours by calling the find contours function. Then, by iterating through them and computing their areas, the largest one can be identified. Next, the spatial moments are computed to find the largest contour's centroid. Finally, a static offset is added to the centroid in the x-axis to approximate the center of the lane as seen in Fig.~\ref{fig3} (on the left of the line if driving in the right lane). While straightforward and computationally efficient, this approach depends on a single line and can be misled by the wrong contour (e.g., opposing lane lines, markings on curves, or other white objects).

\subsubsection{Birds-eye-view with Least Square Regression line fitting}
\begin{figure}[h]
\centerline{\includegraphics[width=0.95\linewidth]{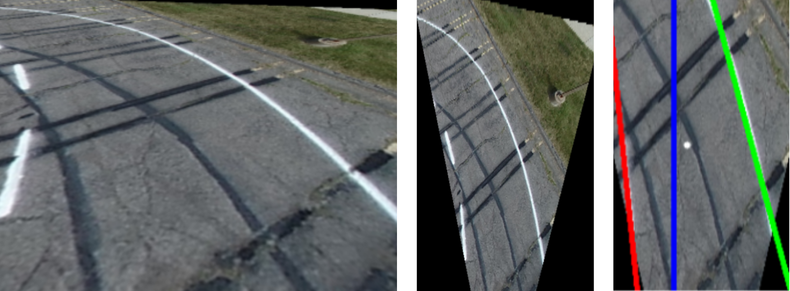}}
\caption{Birds-eye-view transformation with fitted lane lines.}
\label{fig4}
\end{figure}

This method applies a perspective transform to straighten lane lines (Fig~\ref{fig4}), enabling the use of a simple linear model. After Canny edge detection and Hough Transform, lines are filtered to ensure a minimum length and feasible slope (e.g., discarding horizontal lines). The image is then divided into left and right halves; the resulting points are fitted with separate lane-lines using the least squares method, yielding $y = mx + b$ where the slope $m$ and $y$-intercept $b$ are calculated as seen in Eqn.~\ref{eqn1}.

\begin{equation}
    m = \frac{n \sum xy - \sum x \sum y}{n \sum x^2 - (\sum x)^2}, \quad
    b = \frac{\sum y - m \sum x}{n}
\label{eqn1}
\end{equation}

The two line equations define the lane boundaries, and the midpoint is taken as the center. While more robust than a single-line approach, outliers may still impact the fitted lines. Dynamic ROIs, slope thresholds, and fallback logic (e.g., assuming a line at the image edge when undetected) can mitigate noise.

\subsubsection{Linear Lane Search with K-Means}
\begin{figure}[h]
\centerline{\includegraphics[width=0.95\linewidth]{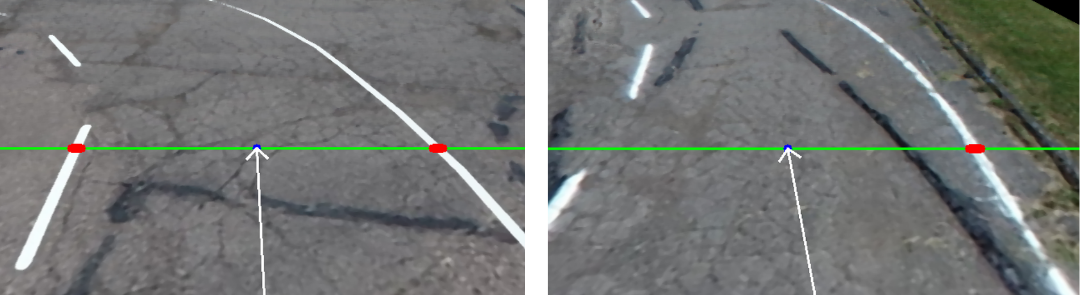}}
\caption{Visualization of Linear Lane Search with K-Means.}
\label{fig5}
\end{figure}

In this approach, only a single horizontal band (row) is analyzed for white pixels. Ideally, two clusters correspond to the left and right lane lines. K-Means with $K=2$ is used to find centroids for each cluster, and these centroids are averaged to determine the lane center\cite{IKOTUN2023178}.  If one or both lane lines are missing, the method reuses the previous frame's centroids (Fig~\ref{fig5}). This algorithm is computationally lightweight and stable; however, extraneous white features near the search row can degrade performance if not filtered out using techniques like histograms and thresholding. 

\subsubsection{Lane Line classification using DBSCAN}
DBSCAN separates dense groups of points from sparse areas and outliers, making it a good fit for lane line clustering. After extracting points via Canny and Hough lines, DBSCAN is applied, forming clusters where points lie within a predefined radius $\epsilon$~\cite{9356727}. The two clusters closest to the bottom of the image (i.e., directly in front of the vehicle) are selected if they exceed $minPoints$, and their centroids are averaged for the lane center. Determining an appropriate $\epsilon$ is crucial, especially for curved lanes. While a bird's-eye transform can help space out converging lines, excessive extension of lines risks merging distinct lane markers.

\begin{figure}[h]
\centerline{\includegraphics[width=0.95\linewidth]{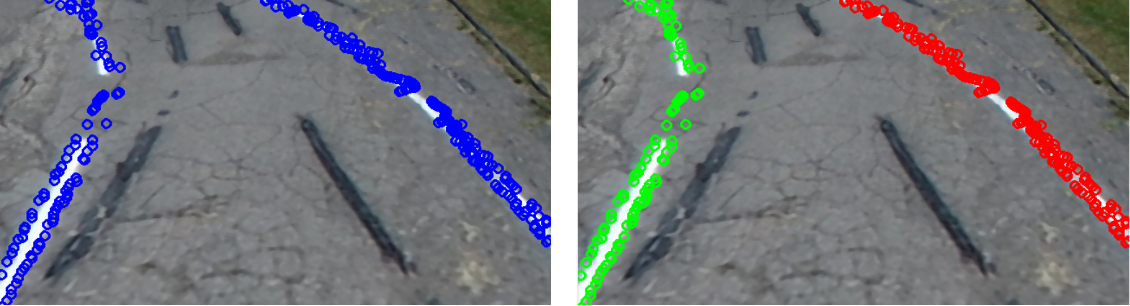}}
\caption{DBSCAN Clustering Lane Lines.}
\label{fig7}
\end{figure}

Density-based clustering is effective for lane line detection because it can handle well-separated lines (Fig.~\ref{fig7}). They can also connect discontinuous segments of the center line, provided an appropriate $\epsilon$. DBSCAN separates lane lines more effectively than simple vertical splits or slope-based methods, which struggle with curved lanes. 

% \begin{figure}[h]
% \centerline{\includegraphics[width=0.65\linewidth]{dbscan.png}}
% \caption{DBSCAN Algorithm.}
% \label{fig6}
% \end{figure}

\subsubsection{DeepLSD Lane Detection}
\begin{figure}[h]
\centerline{\includegraphics[width=0.95\linewidth]{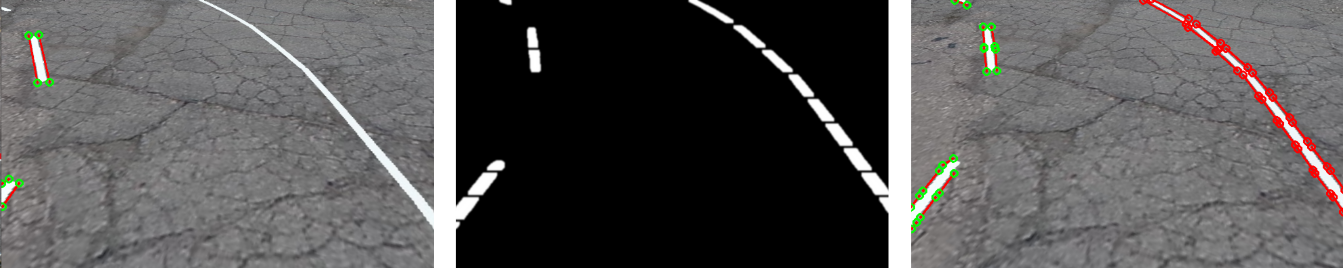}}
\caption{Curve Detection Using DeepLSD.}
\label{fig8}
\end{figure}

This final approach leverages a deep learning-based line detector, DeepLSD~\cite{10204382}, as a performance benchmark. DeepLSD was chosen for its ability to generalize to our test course without retraining, relative to other open-source lane detectors.  It performed well under ideal conditions but it also struggled with identifying lines on the test course. The model detected noise, such as cracks and potholes, and had difficulty with curved lines. Applying the same masking and filtering steps used in traditional pipelines improved performance. To address curve detection, horizontal lines were drawn into the image, converting the curved lines into short straight segments for better analysis (Fig.~\ref{fig8}). 

Detected line segments are filtered by slope and length, and a clustering step (e.g., DBSCAN) can further refine the lane boundary estimate. On the test laptops, inference took approximately 0.15s, necessitating parallelization to maintain the drive-by-wire system's required 50Hz keepalive heartbeat. The coordinates of the centroid between the lane lines are stored globally and updated once the model finishes processing, published to the vehicle during each callback.

\section{Lane Following Algorithms}
\begin{figure}[h]
\centerline{\fbox{\includegraphics[width=0.95\linewidth]{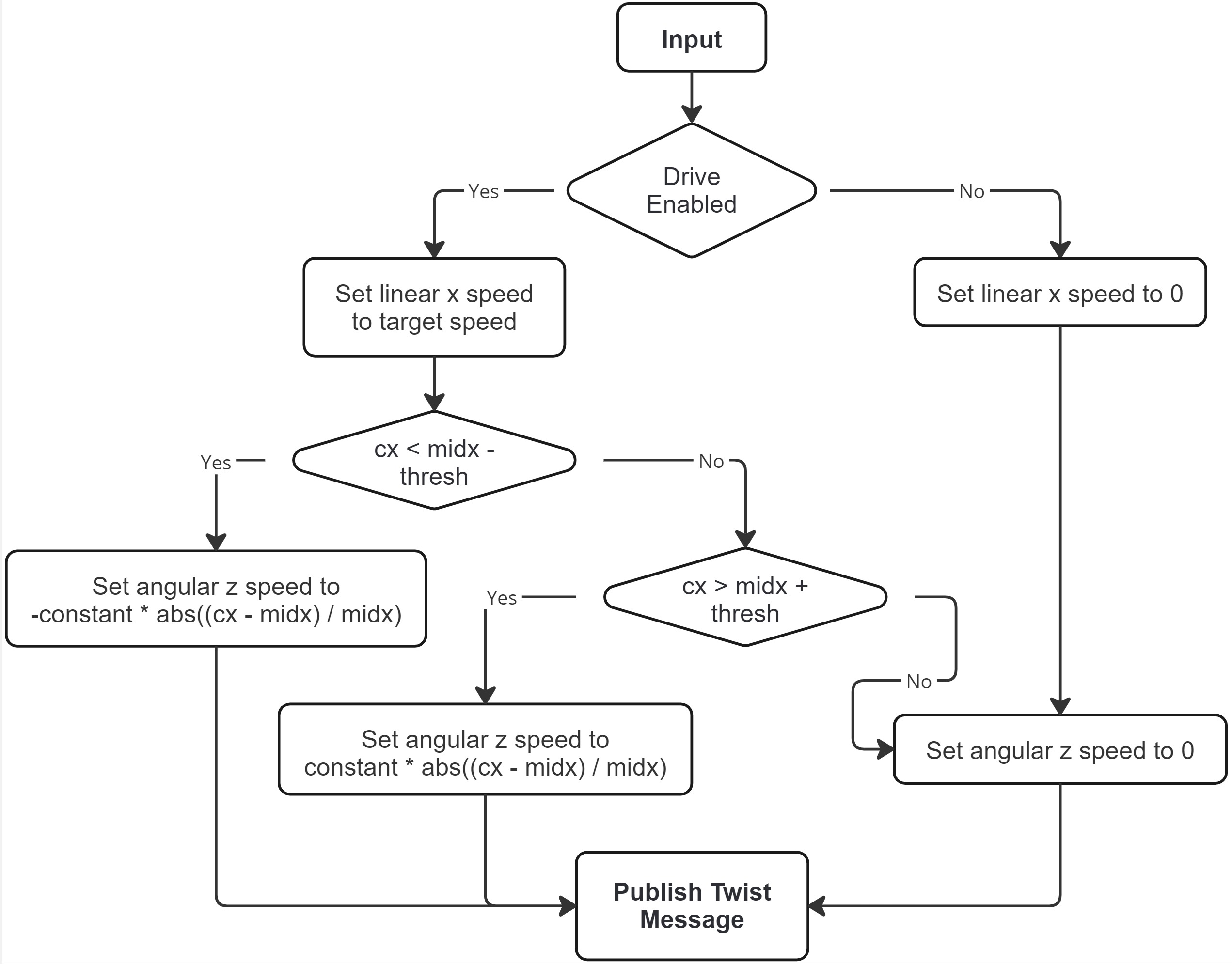}}}
\caption{Lane Centering Using Yaw Rate.}
\label{fig9}
\end{figure}

Once lane lines are detected, the next step is to issue motion commands that center the vehicle within the lane. Two primary methods were used in the vehicle Controller node. The first approach uses ROS Twist messages, which control the vehicle's motion with linear speed along the x-axis (measured in meters per second) and angular velocity, or yaw rate (Fig.~\ref{fig9}), along the z-axis (measured in radians per second). These values are published to the vehicle's \(vel\_cmd\) topic. As an input, the algorithm only needs the offset between the center of the image (denoted as midx), and the center of the lane (denoted as cx) computed by the lane detection algorithms.

\begin{figure}[h]
\centerline{\fbox{\includegraphics[width=0.95\linewidth]{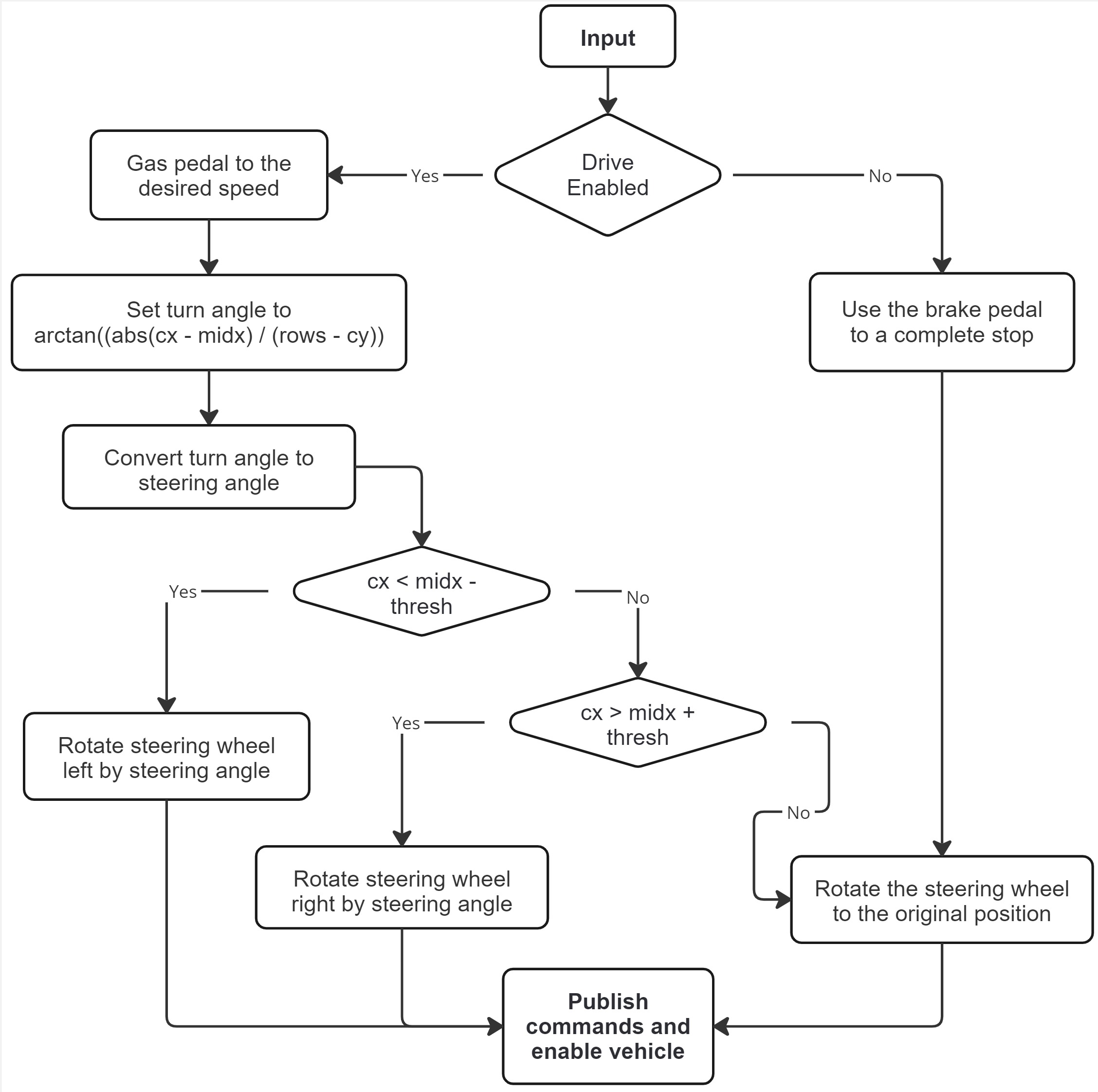}}}
\caption{Lane Centering with DBW Native Commands.}
\label{fig10}
\end{figure}

This approach works at slow speeds but fails to maintain occupant perceived comfort (self-reported) at higher speeds due to the complex kinematics involved. To address these issues, an alternative control method is introduced for the Ackermann steering vehicle \cite{8981710}. Instead of relying on yaw rate control, this method controls the vehicle's steering and pedals interfaces directly, taking advantage of internal nonlinearities. The Dataspeed drive-by-wire system uses custom ROS messages for this purpose. Actuator Messages (SteeringCmd) handle steering, while Unified Control Messages (UlcCmd) manage the pedals (Fig.~\ref{fig10}). 

A key improvement with this method is incorporating the y-offset (denoted as cy) of the computed lane center. Along with cx, midx, and the image height, a turning angle relative to the y-axis is calculated. This turning angle is then converted into a steering angle and transmitted via the SteeringCmd message. Once enabled, these commands result in an enhanced self-reported comfort and lane centering. 

\section{Experiment and Results}
The performance of the five lane detection algorithms was evaluated on the Lot H course. The objective was to complete five consecutive laps on both inner and outer lanes, driving on the right side. DBW Native Commands were employed for lane-following. The outer lane, measuring 97.54 meters long, was driven at a constant 2m/s, while the inner lane, measuring 78.67m was driven at 1.5m/s. These speeds were determined experimentally to be both reliable and comfortable. In the inner lane, there is a sharp turn with a radius of just 4 meters where the lane-following algorithms fail more frequently. In cases where an algorithm failed to complete all five laps, it was re-run at reduced speed, as necessary. Experiments were conducted over several days under varying weather conditions (sunny to cloudy), showcasing the adaptability of each algorithm to environmental changes.

All five algorithms successfully achieved the set goal. Because the speed was held constant across algorithms, the average time showed minimal variation (Table~\ref{tab:counts}). The number of attempts required to complete all laps provides a more telling measure of reliability. The only algorithm to successfully complete all 10 laps on the first attempt was DBSCAN. 

\begin{figure}[h]
\centerline{\includegraphics[width=0.8\linewidth]{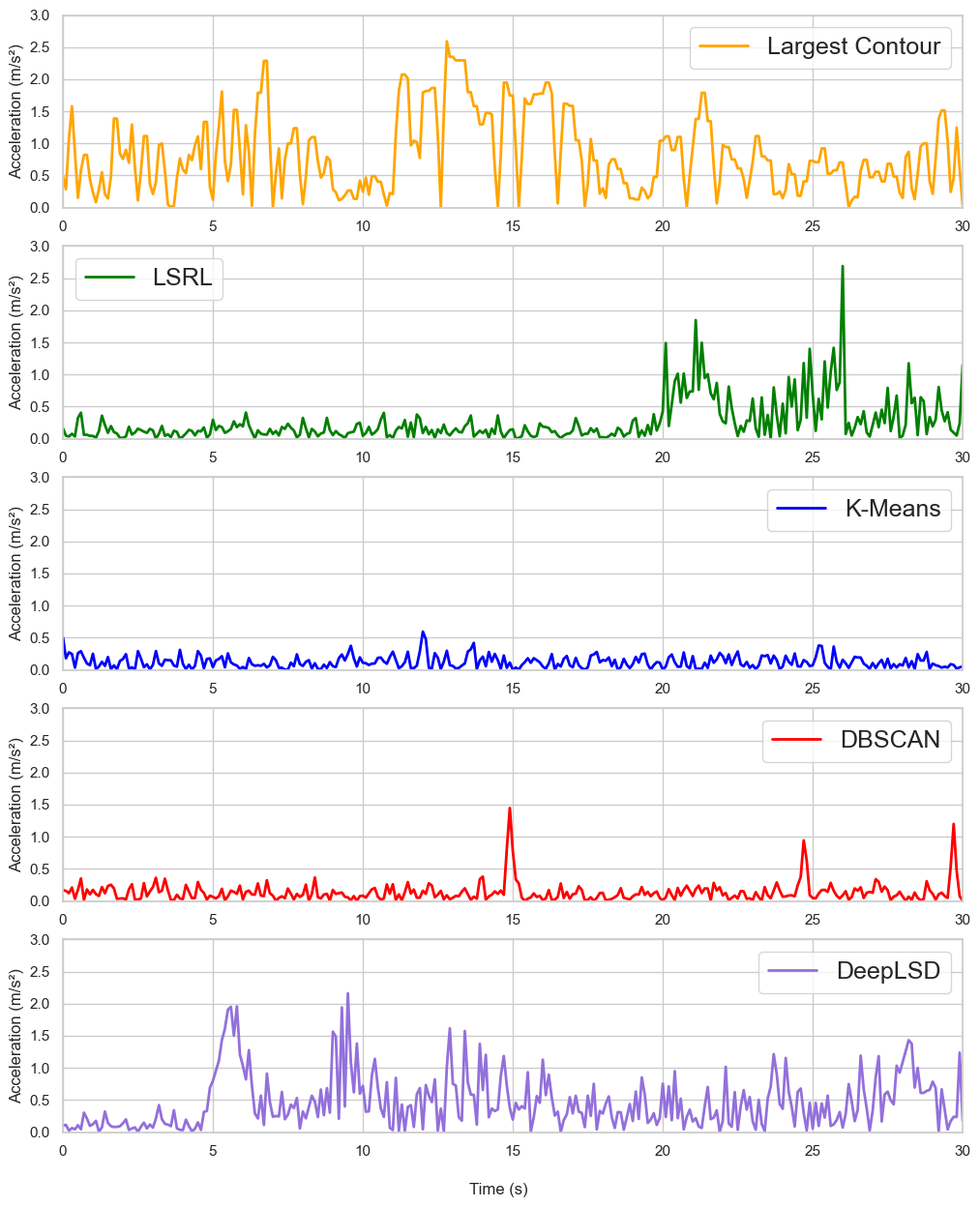}}
\caption{Linear Momentum Measure Plot.}
\label{fig11}
\end{figure}

\begin{table}[h]
    \centering
    \caption{Attempts Needed by Each Algorithm.}
    \footnotesize % Reduce the font size
    \setlength{\tabcolsep}{4pt} % Adjust column separation
    \renewcommand{\arraystretch}{1.2} % Adjust row height
    \begin{tabular}{|l|c|c|c|c|c|}
        \hline
        \textbf{Type} & \textbf{DeepLSD} & \textbf{DBSCAN} & \textbf{K-Means} & \textbf{LSRL} & \textbf{Largest Contour} \\
        \hline
        Inner & 5 & 1 & 2 & 2 & 2 \\
        Outer & 1 & 1 & 1 & 5 & 3 \\
        \hline
        Total & 6 & 2 & 3 & 7 & 5 \\
        \hline
    \end{tabular}
    
    \label{tab:counts}
\end{table}
When speeds were increased, issues with jerkiness and high acceleration during turns became more pronounced. To further investigate these effects, linear and angular momentum were measured using GPS and an Inertial Measurement Unit (IMU).  We focused on the angular-z component, indicating angular momentum. During test runs, GPS coordinates (latitude and longitude) were recorded using Rosbags, with 30-second segments—approximately one lap—extracted for analysis. Linear momentum was calculated by determining the distance between consecutive GPS points using the Haversine formula \cite{1068eb6b-091c-3263-9522-c68d480117c4}. Velocities were computed by dividing distances by the time difference between timestamps, and accelerations were obtained by differentiating velocity with respect to time, as seen in Eqn.~\ref{eqn2} and Fig.~\ref{fig11}.

    \begin{equation}
        v_i = \frac{d_i}{t_i - t_{i-1}}, \quad
        a_i = \frac{v_{i+1} - v_i}{t_{i+1} - t_i}
        \label{eqn2}
    \end{equation}

\begin{figure}[h]
\centerline{\includegraphics[width=0.8\linewidth]{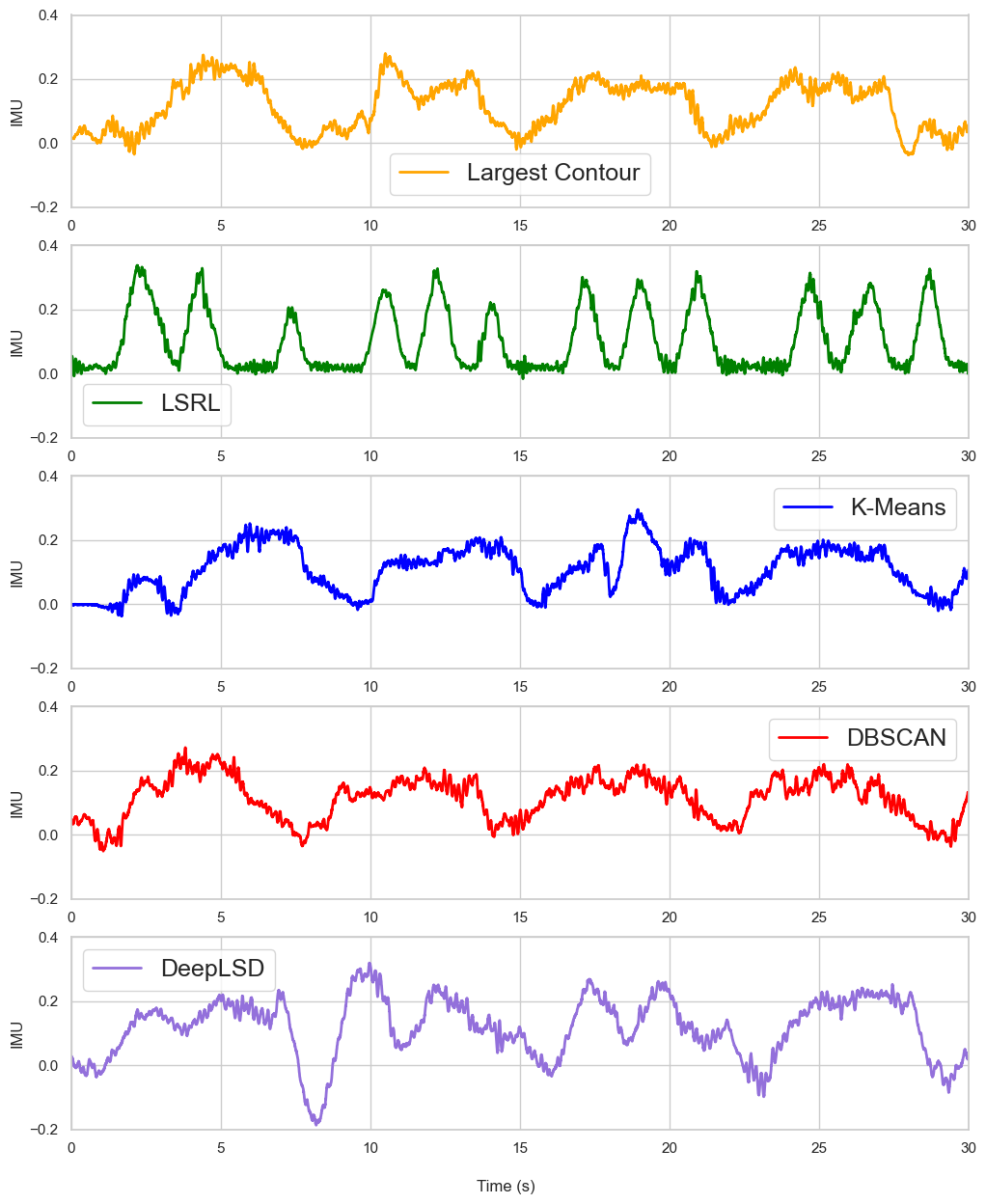}}
\caption{Angular Momentum Measure Plot.}
\label{fig12}
\end{figure}

Fig.~\ref{fig11} shows that the K-Means and DBSCAN-based lane detection algorithms are the best at keeping a steady speed. Sharp turns, inclined slopes, potholes, and the algorithms themselves can cause variations from the target speed. The other three algorithms show heightened instability. In Fig.~\ref{fig12}, the angular momentum plots show the completion of a lap's four turns. Smoother peaks proxy better comfort. Negative peaks indicate that algorithmic overcorrection and recovery, often due to latency. In the LSRL plot, the IMU readings show sharp spikes at various points along each curve, followed by extended periods of remaining at zero. This may be due to the algorithm's lack of foresight, attributable to its reliance on birds-eye projection imagery.

\section{Conclusion and Future Work}

Based on reliability, comfort, speed, and adaptability, the Linear Lane Search with K-Means and the Lane Line Discrimination using DBSCAN emerged as the most robust lane detection algorithms among the five tested.  Both leverage unsupervised learning to effectively discriminate lane lines, enabling the vehicle to maintain a more centered path. These algorithms achieved maximum speeds of 3.5 m/s in the outer lane and 2.5 m/s in the inner lane, with processing times of 10 ms or less per frame, ensuring real-time performance. Their high reliability and low-latency processing highlight their suitability for resource-constrained autonomous vehicles, making them promising candidates for advanced driver-assistance systems and higher levels of vehicle automation. Future work will focus on optimizing lightweight deep learning models for constrained computing resources to achieve performance comparable to unsupervised algorithms such as DBSCAN and K-means.  

\hspace{3cm}
\bibliographystyle{IEEEtran}
\bibliography{refs}
\end{document}